\documentclass[conference]{IEEEtran}
\IEEEoverridecommandlockouts
\usepackage{cite}
\usepackage{amsmath,amssymb,amsfonts}
\usepackage{algorithmic}
\usepackage{graphicx}
\usepackage{textcomp}
\usepackage{xcolor}

\def\BibTeX{{\rm B\kern-.05em{\sc i\kern-.025em b}\kern-.08em
    T\kern-.1667em\lower.7ex\hbox{E}\kern-.125emX}}
\begin{document}

\title{Parkinson’s Disease Freezing of Gait (FoG) Symptom Detection Using Machine Learning from Wearable Sensor Data}

\author{\IEEEauthorblockN{Mahmudul Hasan}
\IEEEauthorblockA{\textit{Department of Computer Science and Engineering} \\
\textit{BRAC University}\\
Dhaka, Bangladesh \\
mahmudul.hasan5@g.bracu.ac.bd}
\and
\IEEEauthorblockN{Dr. Md. Golam Rabiul Alam}
\IEEEauthorblockA{\textit{Department of Computer Science and Engineering} \\
\textit{BRAC University}\\
Dhaka, Bangladesh \\
rabiul.alam@bracu.ac.bd}
}

\maketitle

\begin{abstract}
Freezing of gait (FoG) is a special symptom found in patients with Parkinson’s disease (PD). Patients who have FoG abruptly lose the capacity to walk as they normally would. Accelerometers worn by patients can record movement data during these episodes, and machine learning algorithms can be useful to categorize this information. Thus, the combination may be able to identify FoG in real time. In order to identify FoG events in accelerometer data, we introduce the Transformer Encoder-Bi-LSTM fusion model in this paper. The model’s capability to differentiate between FoG episodes and normal movement was used to evaluate its performance, and on the Kaggle Parkinson's Freezing of Gait dataset, the proposed Transformer Encoder-Bi-LSTM fusion model produced 92.6\% accuracy, 80.9\% F1 score, and 52.06\% in terms of mean average precision. The findings highlight how Deep Learning-based approaches may progress the field of FoG identification and help PD patients receive better treatments and management plans.
\end{abstract}

\begin{IEEEkeywords}
Deep Learning, Time Series Analysis, Parkinson’s Disease, Freezing of Gait, wearable sensors
\end{IEEEkeywords}

\section{Introduction}
The second most widespread neurodegenerative illness and the one with the greatest rate of increase in frequency, associated disability, and mortality is Parkinson’s disease (PD) [1]. 1–2\% of people over 65 have PD, and as the people age, its prevalence is rising quickly [2]. A study of 6620 PD patients revealed that 28\% of them reported FoG every day, and 47\% reported regular FoG [3]. It usually occurs suddenly and is brief, with the motor system being halted for a few seconds to a few minutes [4]. When experiencing FOG, patients frequently describe that their legs are glued to the ground for no apparent reason [5]. In most situations, patients exhibit a minimal amount of freezing in the hospital, whereas their caretakers claim they freeze extensively at residence [6]. For patients, FOG has significant social and medical repercussions as it lowers quality of life [7], disrupts everyday activities, and is a frequent cause of falls [8]. Therefore, developing methods that can aid in lowering the prevalence of FoG is crucial.

Treatment with pharmaceuticals does not work effectively for FoG. Levodopa (LD) is the most often prescribed medication for PD patients’ motor symptoms. The duration of LD’s effects on parkinsonism symptoms ranges from two to six hours, and they gradually fade off. Some individuals experience a slow decline in motor function as a result of this wearing-off effect, whereas others experience a sudden and rather severe decline. It is possible to identify distinct ON and OFF phases for these people, where ON periods denote when the drug is working and OFF periods denote when it isn’t. The right duration of each medication dose decreases with the progression of the disease, necessitating more frequent LD administration [9]. Therefore, in order to alleviate symptoms and enhance mobility, functional non-pharmacologic therapy must be created. 

Mental health issues may have a significant influence in the pathophysiology of FoG, which is a known fact that has not yet been fully utilized [10]. Stress and anxiety are associated with and probably contribute to the incidence of FoG [11]. When a person feels a strong need to act quickly, such as when they have little time to get on an elevator before the doors shut, the likelihood of freezing may become more noticeable. Therefore, by utilizing wearable sensory device technology, it is important to track patient gait and obtain an accurate estimate of the occurrence and intensity of freezing episodes experienced by a patient in their usual environment.

Utilizing ML techniques, the aim of this research is to identify the beginning and end of each
freezing episode, besides the incidence of the following three categories of freezing gait events-Walking, Turning, and Start Hesitation. The aims of this study include: detecting the start and stop of each freezing episode and the occurrence in a series of three types of FoG events, improving the ability of medical professionals to optimally evaluate, monitor, and ultimately, prevent FoG events, participating in the creation of novel techniques and resources for the early identification and monitoring of cognitive and motor decline, and helping researchers better understand when and why FoG episodes occur.

In this work, we design and develop a machine learning (ML) model to detect and classify FoG in patients with PD using wearable sensor data. When wearable sensor devices are used with ML methods, the accuracy of detecting FOG from a lower back accelerometer is relatively high. The key contributions of this research are as follows: 
\begin{itemize}
  \item We proposed a Transformer Encoder-Bi-LSTM fusion model to detect freezing of gait where Transformer Encoder was utilized for feature extraction and Bi-LSTM for classifying the type of FoG event. 
  \item We have tested our model on two benchmark datasets. The proposed model outperformed the state-of-the-art models without extensive data preprocessing and ensemble multiple models.
\end{itemize}

This paper's remaining sections are organized as follows: A thorough analysis of the pertinent literature is given in Section 2. The approach used is explained in detail in Section 3, which also covers the dataset, data processing procedures, the suggested model architecture, and the model training procedure. The experimental results and performance evaluation metrics are presented in Section 4, and an overview of the findings and possible future approaches is provided in Section 5.

\section{Related Work}
Early attempts at identifying FoG episodes relied on the concept of acceleration signals. Progressive work by Han et al. in 2003 utilized a 4-level Daubechies wavelet transform on acceleration data from PD patients’ ankles [12]. They discovered a distinct shift in the main frequency of the signal during freezing events, rising from 2Hz during normal movement to a range of 6 to 8 Hz. Bächlin et al. further refined detection in 2010 by incorporating an additional criterion that differentiated standing from freezing based on the overall energy disparity in the signal [13].

However, these accelerometer-based methods had limitations. They primarily identified freezing patterns resembling tremors, which while common, don’t encompass all FoG presentations. Recognizing this, researchers explored incorporating additional sensors and more sophisticated classifiers. Coste et al. in 2014 demonstrated the limitations of relying solely on acceleration energy [14]. They proposed a new benchmark by utilizing a shank-mounted sensor to measure stride length and cadence alongside acceleration. These additional gait characteristics, combined with a threshold classifier, improved FoG detection accuracy. Another approach involved combining different sensor modalities. Cole et al. in 2011 utilized a two-stage classification system. First, a linear classifier distinguishes between sitting/lying and standing. Then, a Dynamic Neural Network classifier identified freezing during movement. Mazilu et al. further emphasized the value of multimodal approaches, suggesting that incorporating more information could enhance detection speed and accuracy [15]. Cappeci et al. [16] echoed this sentiment, demonstrating improved results when integrating gait factors into deep learning models for FoG detection. For a more comprehensive exploration of FoG identification research, refer to [17].

Several studies explored effective machine learning techniques for FoG detection in real-world settings. Ahlrichs et al [18] introduced a procedure for monitoring FoG occurrences in home environments. They extracted various frequency-based features from 3.2-second data segments. Analysis using a SVM classifier with 10-fold cross-validation yielded impressive results, achieving a sensitivity of 92.3\% and a perfect specificity (100\%). Rodríguez-Martín et al. [19] explored FoG detection in daily activities using an SVM classifier with 55 features encompassing statistical and spectral characteristics. Their LOSO evaluation yielded a sensitivity of 0.7903 and a specificity of 0.7467. Samà et al. [20] further optimized this approach by lessening the number of attributes while maintaining high performance (sensitivity of 91.81\% and specificity of 87.45\%).

The development of accurate FoG detection models is becoming more and more popular with the introduction of wearable sensors and the progress made in ML. Wearable sensors that track gait objectively and continuously include magnetometers, gyroscopes, and accelerometers. Numerous studies have leveraged data from these sensors to develop algorithms for FoG detection. Mancini et al. (2012) demonstrated the feasibility of using accelerometer data to distinguish FoG episodes from normal walking in PD patients [21]. Similarly, Moore et al. (2008) highlighted the potential of wearable sensors to monitor gait and predict FoG, underscoring the need for accurate and real-time detection methods [22]. Traditionally used machine learning techniques including SVM, KNN, and Decision Trees were used in the early research on FoG identification. For instance, Rodríguez-Martín et al. (2017) used SVM to classify FoG episodes based on features extracted from accelerometer data, achieving promising results. However, these models often require extensive feature engineering and may not capture the temporal dependencies in gait data effectively [23]. Recurrent Neural Networks (RNNs) are useful for FoG detection because they can record temporal dependencies. An extension of LSTM called Bi-LSTM analyzes data both forward and backward to give a thorough grasp of the temporal context. For example, Hannink et al. (2017) utilized an LSTM network to classify different gait patterns, including FoG, using wearable sensor data [24]. 

Machine vision is an evaluative technology that uses a machine to see instead of a human. Machine vision-based approaches are more objective than sensor wearables because they don’t require the patient to wear a gadget and don’t interfere with their movement or cause discomfort. A few studies have used the Kinect depth camera, which is a 3D motion capture system, to extract motion data for the purpose of identifying gait abnormalities [25]. However, these cameras need pricey specialized equipment like the Microsoft Kinect. Using RGB technology for 2D key point recognition, as exemplified by OpenPose [26], is an additional technique that may estimate the joint coordinates of individuals in films captured with a monocular camera without the need for external scales or markers. In order to identify FOG using 2D keypoint estimation, a study suggested a unique design for a graph convolutional neural network and obtained good detection performance [27].

\section{Methodology}
Finding the beginning and end of each freezing episode also the recurrence of these
three types of FoG events are the goals of the proposed Parkinson’s FoG detection
model: walking, turning, and start hesitation. To do this, the model has to be
designed with a procedure that receives as input lower-back 3D accelerometer data
from participants, processes the data in a methodical manner, detects the beginning and end of each FoG episode, and generates predictions of three folds. The model uses the input data to detect freezing of
gait events after it has undergone preprocessing. Since the distribution of the two
datasets differs, it is required to create two distinct models: one for tdcsfog and one
for defog. The proposed FoG detection system is composed of several phases:
(1) Collection of data, (2) Processing the input data, (3) Propose FoG detection
Model, (4) Classification of FoG into three types of events. Figure 1 displays a high-level overview of the suggested FoG detection system.

\begin{figure}[htbp]
\centerline{\includegraphics[width=0.5\textwidth]{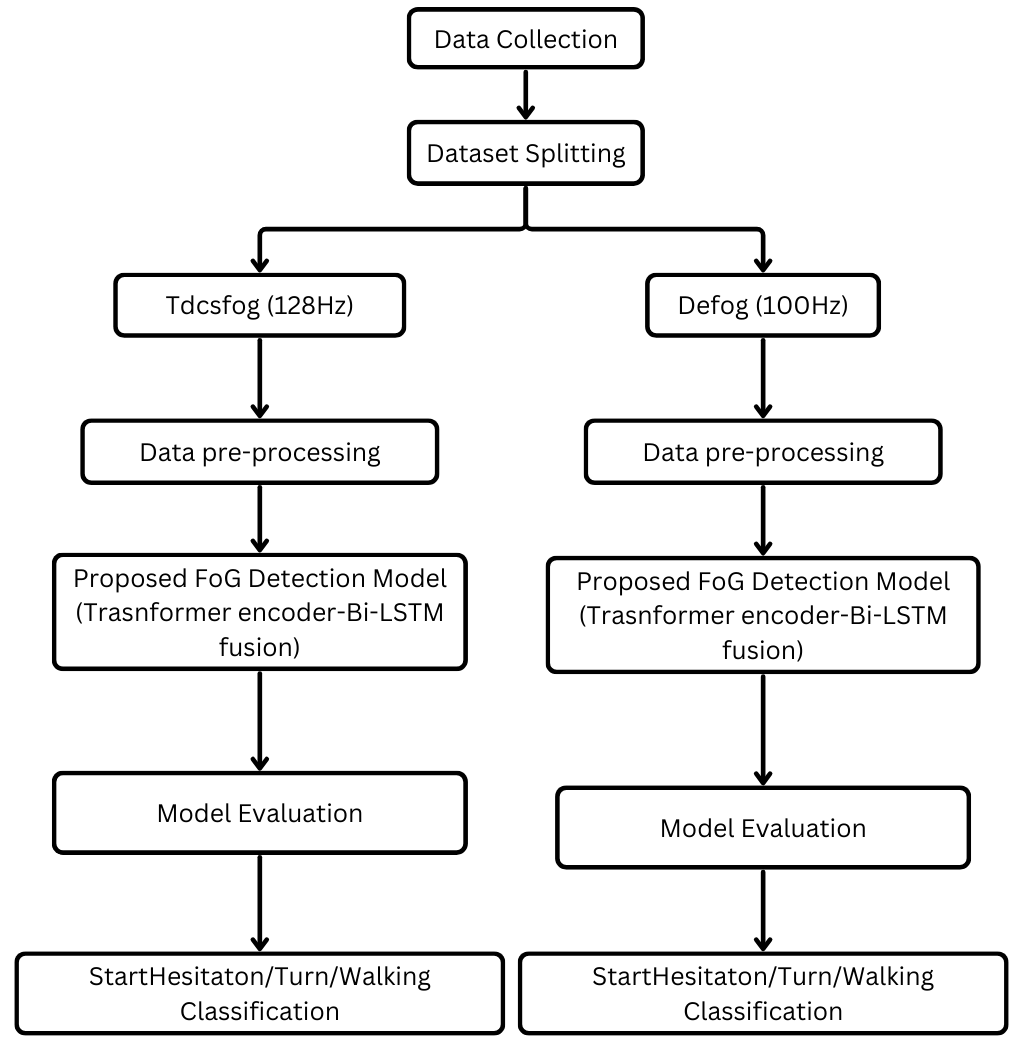}}
\caption{Top Level Overview of the Proposed FoG Detection System}
\label{fig1}
\end{figure}

In the proposed FoG
detection model, two data sets (explained in Section 3(A)) have been used. The
next phase is the data processing steps described in Section 3(B). After that, we
proposed a Transformer Encoder-Bi-LSTM fusion model (explained in Section 3(C))
to detect freezing episodes from the data and categorize them into three discrete
classes. Finally, the model was evaluated by
the mean average precision of predictions for each event class.

\subsection{Dataset}
Machine learning based FoG classification has been conducted on two freezing of gait datasets, namely tdcsfog and defog. These data sets have been acquired from publicly accessible database available on Kaggle [28]. The expert reviewers annotated these data sets, which documented the freezing of gait episodes. 

\subsubsection{The TdcsFOG Dataset (tdcsfog)}
Consisting of data series collected in the lab. During each visit, individuals wore a 3D accelerometer on their lower back. Every trial that caused FoG was captured on camera and examined offline. Before the test protocol begins, there is a brief (2–3 s) interval of silent standing while data recordings.
\subsubsection{The DeFOG Dataset (defog)}
 Consisting of data series gathered while subjects were completing a program designed to induce FOG in their homes. Two trips to the subject's home surroundings were part of this investigation. The subjects were assessed at both the off and on medication states at each visit. The subjects wore a 3D accelerometer on their lower backs during the motor evaluation, which gathered data.

The following protocols described by Ziegler et al. 2010 [29] shown in Figure 2 were performed each time for collecting data:
\begin{enumerate}
  \item The 4-Meter Walk Test: Involves the participant walking a distance of 4 meters at their usual, comfortable pace. The time needed to complete the walk is collected using a stopwatch, and the test may be repeated multiple times, with the average time used for analysis.
  \item The Timed Up \& Go (TUG) Single Task: It requires the participant to being seated in a standard chair. Upon the command ``go", the participant stands up, walks a distance of 3 meters, turns around, walks back, and sits down again. The total time needed to complete the task is recorded, and the test assesses mobility, balance, and functional ability.
  \item The Timed Up \& Go (TUG) Dual Task: The participant performs the standard TUG test while simultaneously being asked to subtract a specified number out loud while walking. The total time needed to complete the task and the accuracy of the subtraction are recorded to evaluate the effect of cognitive load on mobility.
  \item The Turning - Single Task: Involves the participant performing four 360-degree turns, alternating the direction with each turn. The participant’s speed, stability, and turning technique are observed and recorded.
  \item The Turning – Dual Task: The participant performs the same turning task as before, but with an added cognitive challenge of subtracting a specified number while turning. The time, stability, and accuracy of the cognitive task are recorded.
  \item The Hotspot Door Test: Involves the participant walking to a designated door, opening it, entering an adjacent room, turning around, and returning to the starting point. This test simulates real-life mobility challenges, such as navigating through doorways and making directional changes. The total time needed and any hesitations or difficulties, such as freezing of gait, are recorded.
  \item The Personalized Hotspot Test: The participant is asked to identify an area within their home, that typically triggers freezing of gait (FoG). The participant then walks through this area under observation, with the occurrence and duration of FoG episodes noted, along with any compensatory strategies used. 
\end{enumerate}

\begin{figure}[htbp]
\centerline{\includegraphics[width=0.4\textwidth]{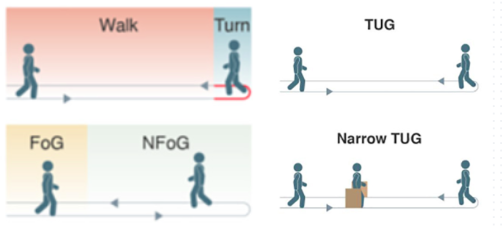}}
\caption{Data collection protocols [30]}
\label{fig2}
\end{figure}

Field description of tdcsfog and defog dataset:
\begin{itemize}
    \item Time: An integer timestep. The datasets differ in their sampling rates: tdcsfog data is collected at 128Hz, whereas defog data is recorded at 100Hz.
    \item Start Hesitation, Turn, Walking: Variables that indicate the occurrence of each event kind.
    \item AccV, AccML, and AccAP: Figure 3 illustrates the three axes of acceleration from a lower-back sensor: vertical (V), mediolateral (ML), Aanteroposterior (AP). For tdcsfog and defog, the data is expressed in m/s² and g, respectively.
\end{itemize}

\begin{figure}[htbp]
\centerline{\includegraphics[width=0.5\textwidth]{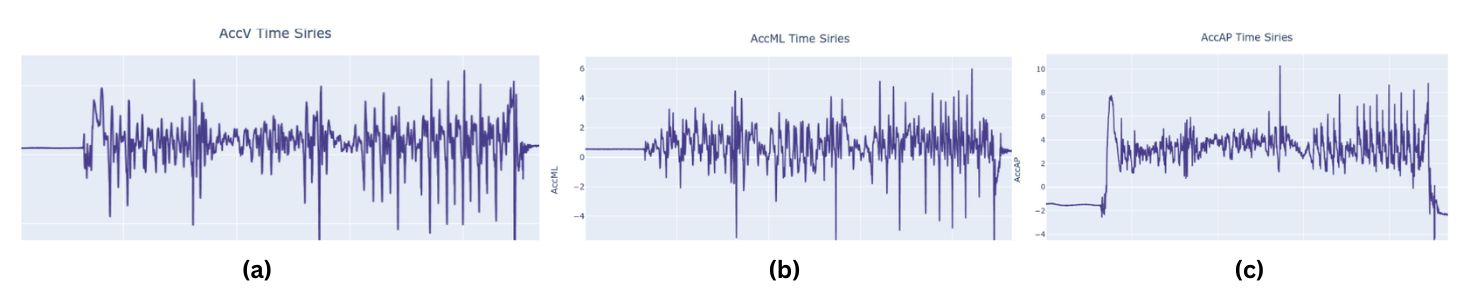}}
\caption{Time series features: (a) AccV, (b) AccML, and (c) AccAP}
\label{fig3}
\end{figure}

A detailed statistical analysis of the dataset provides valuable insights. Figure 4(a) illustrates the distribution of FoG episodes across different events: the majority (79.8\%) occurred during turn events, 15.9\% while walking, and 4.28\% during the Start Hesitation stage. Additionally, Figure 4(b) presents a histogram of FoG duration, showing that most episodes lasted 5 seconds or less, highlighting the brief but critical nature of these occurrences. 
\begin{figure}[htbp]
\centerline{\includegraphics[width=0.5\textwidth]{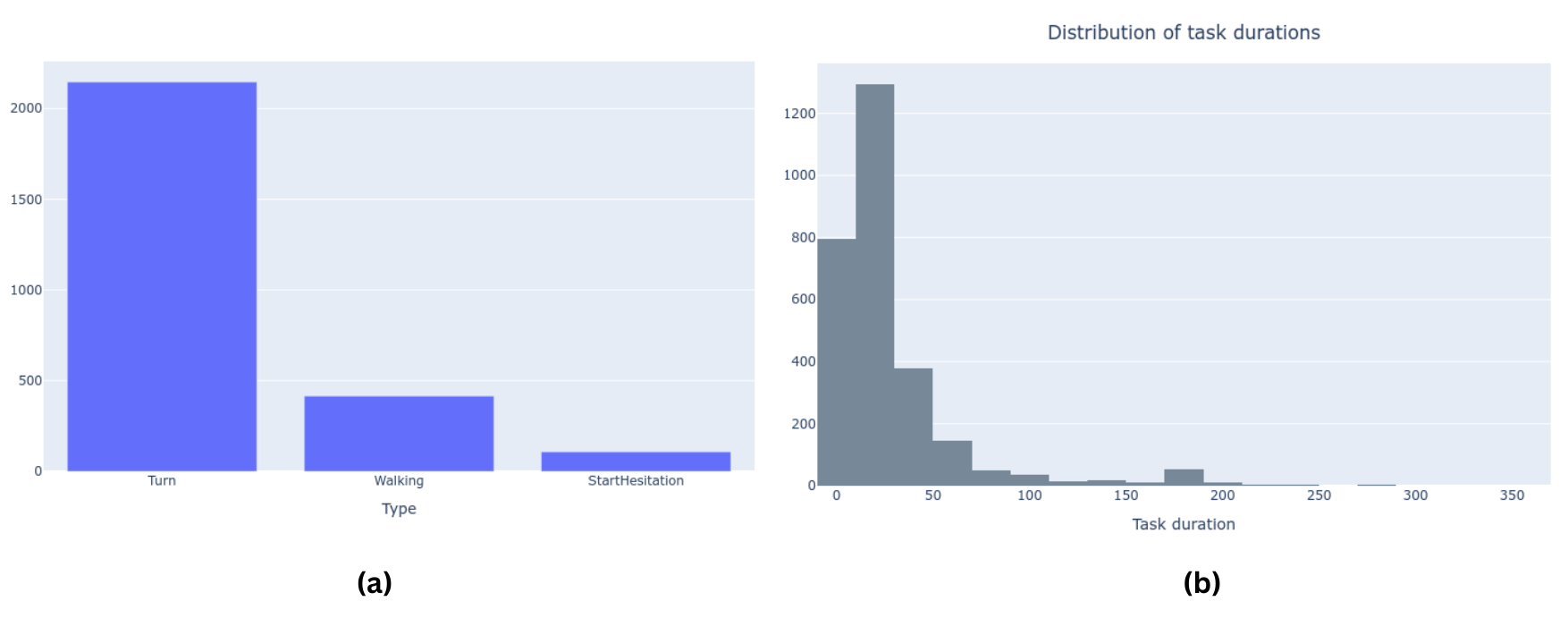}}
\caption{FoG statistics: (a) distribution of FoG events, and (b) FoG-duration distribution}
\label{fig4}
\end{figure}

\subsection{Data Pre-processing}
\subsubsection{Normalization}
Mean-std normalization is applied to the acceleration columns : AccV, AccML, and AccAP for both tdcsfog and defog data series. The differences in measurement units for tdcsfog and defog dataset can lead to inconsistencies when the data is processed. Mean-std normalization standardizes these values, ensuring that the different units do not affect the modeling process. Moreover, the datasets have different sampling frequencies. Normalizing the features ensure that the variance due to different sampling rates is minimized, providing a uniform scale across the datasets. 

\subsubsection{Reducing the Sampling Rate of the Data}
In this work, the frequency of the target data is reduced to improve the model’s performance. The original target data has a high resolution to capture detailed, high-frequency information over time. While such high-resolution data can be useful, deep learning models often struggle to handle targets with a complex structure or excessive detail. To address this, the target data, initially in a high-resolution format, is reshaped and reduced through a series of transformations. The first step reshapes the target data into smaller patches, reducing the temporal resolution. The data is then transposed to facilitate operations across specific dimensions. Lastly, by choosing the most important value within each patch, the TensorFlow reduction max operation is applied throughout the patch dimension, reducing the complexity of the target data. This reduction in resolution simplifies the structure of the target data, enabling the model to better focus on the core signals and features that are relevant for learning.

\subsubsection{Partitioning Data into Fixed-length Blocks}
By copying and selecting specific columns, and then converting these into a standardized numerical format, the method prepares the data for further analysis using patches like Vision Transformers. Padding the series to make its length a multiple of the block size ensures that all blocks are uniformly sized, which is crucial for model compatibility. The use of overlapping blocks, determined by the block stride, allows the method to capture sequential dependencies and transitional features within the series, enhancing the model’s ability to detect patterns like FoG events. By extracting these blocks as individual units containing start and end indices along with the values, the processing facilitates the efficient transformation of continuous time series data into a format suitable for training and testing predictive models. 

\subsection{Proposed Transformer encoder-Bi-LSTM Fusion Model}
In this work, a Transformer Encoder and two Bi-LSTM layers were combined to create a hybrid deep learning model that can identify FoG episodes shown in Figure 5. To find long-range relationships in the data, the model first splits the accelerometer time-series data into blocks, which are then processed by a Transformer Encoder that employs multi-head self-attention. Each encoder layer applies multi-head attention with residual connections and layer normalization, followed by a feed-forward network with dropout to enhance generalization. To incorporate temporal positional information, a trainable position encoding is added to the input, which allows the Transformer to capture the sequential structure of the data. After passing through multiple Transformer encoder layers, the features are further processed by two Bi-LSTM layers, which model both forward and backward dependencies in the time-series data, crucial for identifying transitions between FOG episodes. Finally, the model outputs the confidence score of three FOG event types (Start Hesitation, Turn, and Walking) using a dense layer followed by a sigmoid activation function.
\begin{figure}[htbp]
\centerline{\includegraphics[width=0.5\textwidth]{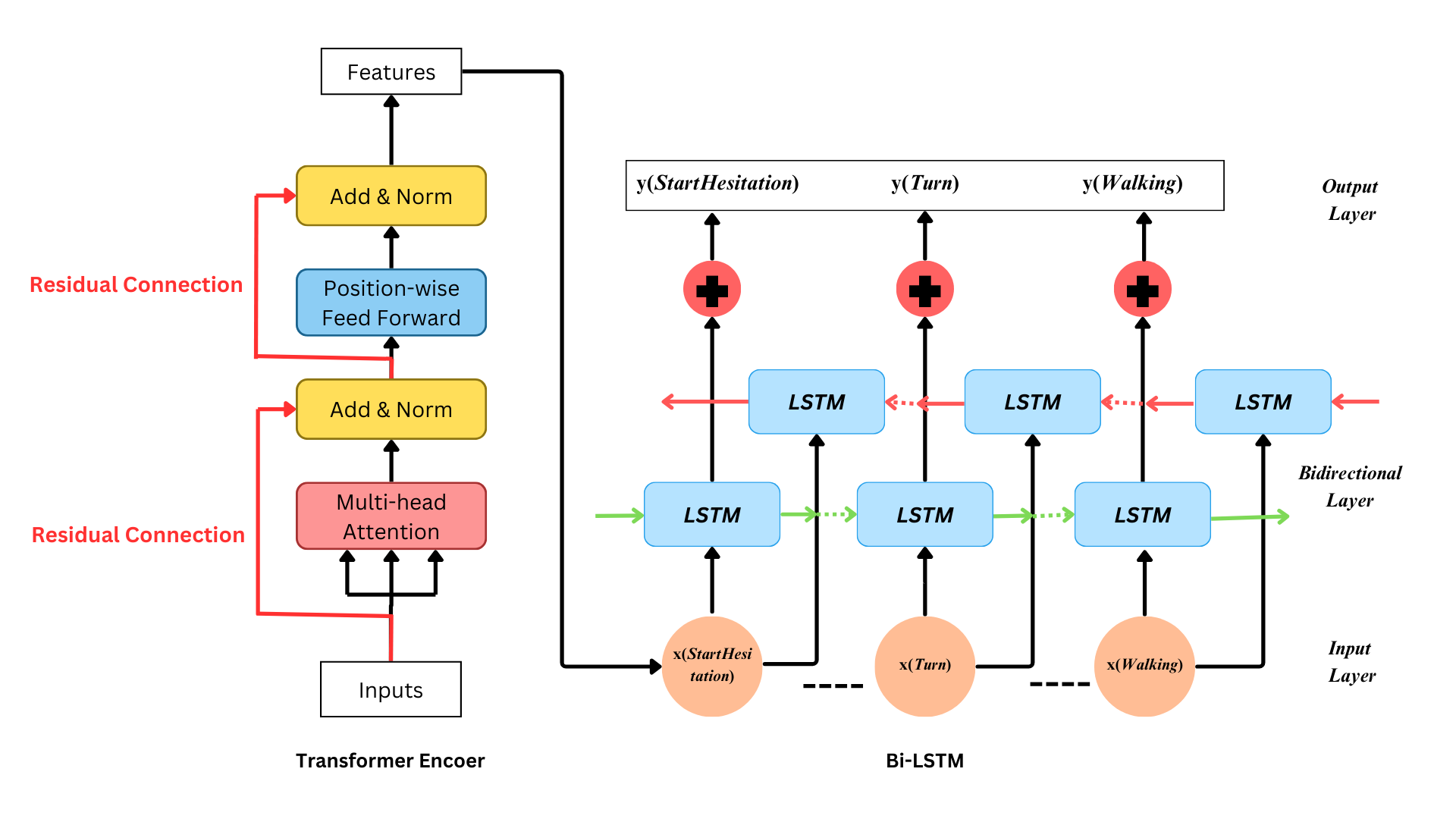}}
\caption{Proposed Transformer Encoder-Bi-LSTM Fusion Model}
\label{fig5}
\end{figure} 

\subsection{Model Training}
In this study, several key hyperparameters were fine-tuned to optimize the detection and classification of FoG episodes using Transformer Encoder-Bi-LSTM fusion model. All the training was conducted with TensorFlow capabilities and Kaggle Notebook resources. To prevent overfitting and improve generalization, dropout is applied at multiple stages, including the first layer and the Transformer Encoder. A key aspect of the training process is the learning rate schedule, which starts with a warm-up phase to stabilize initial training and then maintains a gradual learning rate for fine-tuning. Batch size and the number of steps per epoch are optimized to balance computational efficiency and model stability, while the model also benefits from accelerated training, allowing it to process the large dataset effectively. Each of these parameters contributes to the model's capacity to precisely detect and distinguish FoG events in real-time. The following parameters were used in the model configuration: \\
\begin{itemize}
  \item Block Size: The size of each input segment processed by the model.  
  \item Patch Size: The size of smaller chunks into which each block is divided. 
  \item Fog Model Dimension: The dimensionality of the feature embeddings.
  \item Attention Head: The number of attention heads in the multi-head self-attention
  mechanism. 
  \item Encoder Layer: The number of encoder layers in the Transformer.
  \item First Dropout: Dropout rate applied to the input or initial layers to prevent overfitting.
  \item Encoder Dropout: Dropout rate applied to the Transformer encoder, enhancing generalization by randomly deactivating neurons during training.
  \item MHA Dropout: Dropout rate applied within the multi-head attention mechanism.
\end{itemize}

The loss function used in this research is based on binary cross-entropy (BCE) that computes the loss between the real values and the model's predicted output without reducing the dimensionality immediately, enabling flexibility in further processing. The function first expands the dimensions of both the real values and the output to match a shape suitable for element-wise comparison. Next, two particular columns of the actual tensor are multiplied to create a mask that represents circumstances or occurrences that should be taken into account when calculating the loss. This mask is expanded and tiled across the target dimensions to match the shape of the loss tensor, ensuring that only relevant parts of the input contribute to the overall loss calculation. After that, all the masked loss values are added, and they are normalized using the total mask sum to provide the average loss value that only represents the valid segments. This approach effectively handles the sparsity of relevant events and ensures the model focuses on learning from meaningful parts of the data, improving the accuracy and robustness of the detection model. 

The optimizer used in this model is a custom implementation of the Adam optimizer from TensorFlow. 

\section{Result and Discussion}
\subsection{Performance Evaluation Metrics}
The performance of the proposed model was assessed using a number of evaluation indicators. 
\begin{itemize}
    \item Mean average precision (mAP): It is the mean score of the average precision scores for each event class. The mean average precision is shown in the Equation 3 where AP\textsubscript{k} is the average precision of class k and n is the number of classes. 

    \begin{equation}
        mAP = \frac{1}{n}\sum_{k=1}^{k=n} AP_k 
    \end{equation}
    \item Accuracy: Accuracy, shown in Equation 8, is the percentage of correct classifications that a trained model achieves.
    \begin{equation}
        Accuracy = \frac{Correct \ Predictions}{All \ Predictions} * 100\%
    \end{equation}
    \item F1 Score: The F1 score, as defined in Equation 9, is the harmonic mean of precision and recall. It strikes a balance between these two criteria, resulting in a single score that indicates both the model's accuracy in detecting true positives and its ability to extract all relevant instances.
    \begin{equation}
        F1-score = \frac{2*Precision*Recall}{Precision \ + \ Recall}
    \end{equation}
\end{itemize}

\subsection{Result}

The results of the local cross-validation for the Tdcsfog training models are summarized in Table 1, showcasing a range of mean Average Precision (mAP) scores from 0.4866 to 0.6159. This range indicates a moderate level of overall predictive performance across the dataset. A closer examination of the Average Precision (AP) scores for the three specific outcomes—StartHesitation, Turn, and Walking—reveals considerable variability in the model's ability to predict each activity type. For StartHesitation, the AP values vary significantly, ranging from 0.3736 to 0.6076, reflecting the challenges in accurately identifying hesitation events at the beginning of movement. In contrast, the Turn outcome demonstrates much higher and more consistent AP values, ranging from 0.8636 to 0.9026, suggesting that the model performs reliably well in predicting turning motions. However, the Walking activity exhibits the lowest and most variable AP scores, with values ranging from 0.2006 to 0.4766, highlighting the difficulty in capturing walking-related features accurately. These results underscore the heterogeneous nature of the model's performance, with notable differences in prediction accuracy across different activity types, which may be influenced by the inherent complexity and variability of the activities themselves. 
 \begin{table}[htbp]
\caption{Training Results of Tdcsfog Model using Transformer endcoder-Bi-LSTM Fusion}
\begin{center}
\begin{tabular}{|c|c|c|c|c|}
\hline
 Config. & StartHesitationAP & TurnAP & WalkingAP & mAP \\
 \hline
 1 & 0.3736 & 0.8856 & 0.2006 & 0.4866 \\
 \hline
 2 & 0.6076 & 0.8636 & 0.2956 & 0.5889  \\
 \hline
 3 & 0.4876 & 0.8926 & 0.4436 & 0.6079 \\
 \hline
 4 & 0.4686 & 0.9026 & 0.4766 & 0.6159 \\
 \hline
\end{tabular}
\end{center}
\end{table} 

This study evaluated the performance of various machine learning architectures for detecting and classifying FoG episodes, with mean average precision (mAP) serving as the primary assessment metric. The proposed Transformer Encoder-Bi-LSTM model achieved a mAP score of 0.5206, surpassing the top three solutions from the Kaggle Freezing of Gait (FoG) Detection Challenge, described by Amit et al. in 2024 [31]. Specifically, the mAP scores of the 1st, 2nd, and 3rd place models in the competition were 0.5140, 0.4509, and 0.4364, respectively, as illustrated in Figure 6. This comparative analysis underscores the potential of the proposed architecture as a competitive and computationally efficient solution for FoG detection and classification.
\begin{figure}[htbp]
\centerline{\includegraphics[width=0.5\textwidth]{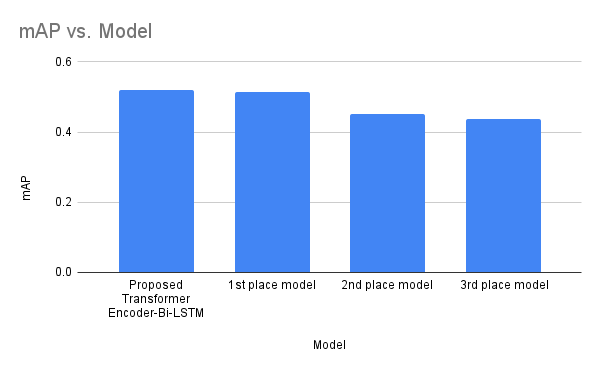}}
\caption{Comparison of Proposed Model with respect to Mean Average Precision}
\label{fig6}
\end{figure}

The performance comparison of the proposed Transformer Encoder-Bi-LSTM fusion model using additional evaluation metrics—precision, recall, F1 score, accuracy, and specificity—is illustrated in Figure 7. The proposed model achieved an F1 score of 0.809, accuracy of 0.926, precision of 0.842, recall of 0.792, and specificity of 0.952. These results are compared against the top three models from the competition, with the Transformer Encoder-Bi-LSTM consistently outperforming them across most evaluation metrics. Notably, the model demonstrates strong recall performance, effectively identifying a high proportion of true positive instances while maintaining robust precision and specificity, showcasing its overall reliability and effectiveness.
\begin{figure}[htbp]
\centerline{\includegraphics[width=0.5\textwidth]{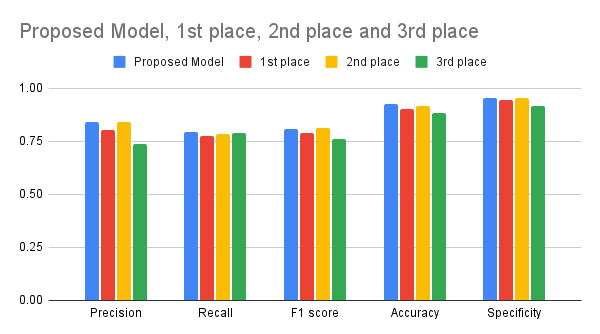}}
\caption{Comparison of Proposed Model with respect to popular evaluation metrics}
\label{fig7}
\end{figure}
\subsection{Discussion}
The results show how to achieve competitive performance in FOG detection without the need for ensembling complex models and extensive data processing. This simplicity not only makes implementation easier, but it also makes it relatively easy for other researchers who are interested to adopt and expand upon the published methodology. In order to improve the standard of life of PD patients, reliable automatic FOG detection is an important aspect to consider. The present study reported a highly ranked solution to the Parkinson’s FoG detection. Results show that using a combination of Transformer Encoder and two Bidirectional LSTM layers can be successfully applied to detect FOG related events in accelerometer data from different settings without applying sophisticated pre-processing.

One key limitation is the computational complexity of the used architectures which make it challenging to deploy the model on edge devices such as wearable sensors, where computational power and battery life are constrained.  Another limitation is related to the real-time performance and latency. Although the model has strong predictive capabilities, ensuring it operates in real-time with minimal lag is crucial for practical applications, such as triggering therapeutic responses for Parkinson's patients during a FoG episode. 

Future work could focus on optimizing the model for real-time deployment. The model could benefit from incorporating data from multiple sensors, such as combining accelerometer data with gyroscope or electromyography inputs, to enhance detection capabilities.

\section{Conclusion}
As freezing of gait can lead to fall and the majority of PD patients are older, therefore falls can cause fractures or even death. Finding ways to diagnose FOG and stop patients from falling is therefore crucial. An effective FOG recognition model based on deep learning was created in this study. This diagnostic examination service lessens the financial burden on PD patients in addition to mitigating the effects of individual subjectivity. Although there are alternative methods, FOG-provoking protocols are used in the majority of FOG evaluation methodologies. Individuals with FOG are captured on camera while engaging in activities that are likely to exacerbate the condition. After watching the footage, experts score each frame to determine when FOG happened. This method of scoring is quite time-consuming and necessitates specialized knowledge, but it is also reasonably sensitive and dependable. Another approach is to use wearable technology to enhance FOG-provoking testing. FOG detection is made easier with additional sensors, but usability and compliance may suffer. Thus, the optimal strategy might be a mix of these two techniques because the accuracy of identifying FOG from a lower back accelerometer is comparatively high when paired with machine learning techniques.


\begin{thebibliography}{00}
\bibitem{b1} E. R. Dorsey, A. Elbaz, E. Nichols, N. Abbasi, F. Abd-Allah, A. Abdelalim, J. C. Adsuar, M. G. Ansha, C. Brayne, J.-Y. J. Choi, et al., “Global, regional, and national burden of parkinson’s disease, 1990–2016: A systematic analysis for the global burden of disease study 2016,” The Lancet Neurology, vol. 17, no. 11, pp. 939–953, 2018.
\bibitem{b2} M. d. De Rijk, L. Launer, K. Berger, M. Breteler, J. Dartigues, M. Baldereschi, L. Fratiglioni, A. Lobo, J. Martinez-Lage, C. Trenkwalder, et al., “Prevalence of parkinson’s disease in europe: A collaborative study of population-based cohorts. neurologic diseases in the elderly research group.,” Neurology, vol. 54, no. 11 Suppl 5, S21–3, 2000.
\bibitem{b3} M. Macht, Y. Kaussner, J. C. Möller, K. Stiasny-Kolster, K. M. Eggert, H.-P. Krüger, and H. Ellgring, “Predictors of freezing in parkinson’s disease: A survey of 6,620 patients,” Movement disorders, vol. 22, no. 7, pp. 953–956, 2007.
\bibitem{b4} J. G. Nutt, B. R. Bloem, N. Giladi, M. Hallett, F. B. Horak, and A. Nieuwboer, “Freezing of gait: Moving forward on a mysterious clinical phenomenon,” The Lancet Neurology, vol. 10, no. 8, pp. 734–744, 2011.
\bibitem{b5} J. Schaafsma, Y. Balash, T. Gurevich, A. Bartels, J. M. Hausdorff, and N. Giladi, “Characterization of freezing of gait subtypes and the response of each to levodopa in parkinson’s disease,” European journal of neurology, vol. 10, no. 4, pp. 391–398, 2003.
\bibitem{b6} A. Nieuwboer, W. d. Weerdt, R. Dom, and E. Lesaffre, “A frequency and correlation analysis of motor deficits in parkinson patients,” Disability and rehabilitation, vol. 20, no. 4, pp. 142–150, 1998.
\bibitem{b7} A. De Boer, W. Wijker, J. Speelman, and J. De Haes, “Quality of life in patients with parkinson’s disease: Development of a questionnaire.,” Journal of Neurology, Neurosurgery \& Psychiatry, vol. 61, no. 1, pp. 70–74, 1996.
\bibitem{b8} B. R. Bloem, J. M. Hausdorff, J. E. Visser, and N. Giladi, “Falls and freezing of gait in parkinson’s disease: A review of two interconnected, episodic phenomena,” Movement disorders: official journal of the Movement Disorder Society, vol. 19, no. 8, pp. 871–884, 2004.
\bibitem{b9} N. Giladi, M. McDermott, S. Fahn, S. Przedborski, J. Jankovic, M. Stern, C. Tanner, and P. S. Group, “Freezing of gait in pd: Prospective assessment in the datatop cohort,” Neurology, vol. 56, no. 12, pp. 1712–1721, 2001.
\bibitem{b10} N. Giladi and J. M. Hausdorff, “The role of mental function in the pathogenesis of freezing of gait in parkinson’s disease,” Journal of the neurological sciences, vol. 248, no. 1-2, pp. 173–176, 2006.
\bibitem{b11} S. Rahman, H. Griffin, N. Quinn, and M. Jahanshahi, “The factors that induce or overcome freezing of gait in parkinson’s disease,” Behavioural neurology, vol. 19, no. 3, pp. 127–136, 2008.
\bibitem{b12} J. H. Han, W. J. Lee, T. B. Ahn, B. S. Jeon, and K. S. Park, “Gait analysis for freezing detection in patients with movement disorder using three dimensional acceleration system,” in Proceedings of the 25th Annual International Conference of the IEEE Engineering in Medicine and Biology Society (IEEE Cat. No. 03CH37439), IEEE, vol. 2, 2003, pp. 1863–1865.
\bibitem{b13} M. Bachlin, M. Plotnik, D. Roggen, I. Maidan, J. M. Hausdorff, N. Giladi,
and G. Troster, “Wearable assistant for parkinson’s disease patients with the freezing of gait symptom,” IEEE Transactions on Information Technology in Biomedicine, vol. 14, no. 2, pp. 436–446, 2009.
\bibitem{b14} C. A. Coste, B. Sijobert, R. Pissard-Gibollet, M. Pasquier, B. Espiau, and C. Geny, “Detection of freezing of gait in parkinson disease: Preliminary results,” Sensors, vol. 14, no. 4, pp. 6819–6827, 2014.
\bibitem{b15} S. Mazilu, M. Hardegger, Z. Zhu, D. Roggen, G. Tröster, M. Plotnik, and J. M. Hausdorff, “Online detection of freezing of gait with smartphones and machine learning techniques,” in 2012 6th International Conference on Pervasive Computing Technologies for Healthcare (PervasiveHealth) and Workshops, IEEE, 2012, pp. 123–130.
\bibitem{16} M. Capecci, L. Pepa, F. Verdini, and M. G. Ceravolo, “A smartphone-based architecture to detect and quantify freezing of gait in parkinson’s disease,” Gait \& posture, vol. 50, pp. 28–33, 2016.
\bibitem{b17} A. L. Silva de Lima, L. J. Evers, T. Hahn, L. Bataille, J. L. Hamilton, M. A. Little, Y. Okuma, B. R. Bloem, and M. J. Faber, “Freezing of gait and fall detection in parkinson’s disease using wearable sensors: A systematic review,” Journal of neurology, vol. 264, pp. 1642–1654, 2017.
\bibitem{b18} C. Ahlrichs, A. Sama, M. Lawo, J. Cabestany, D. Rodriguez-Martin, C. Perez- Lopez, D. Sweeney, L. R. Quinlan, G. O. Laighin, T. Counihan, et al., “Detecting freezing of gait with a tri-axial accelerometer in parkinson’s disease patients,” Medical \& biological engineering \& computing, vol. 54, pp. 223–233, 2016.
\bibitem{b19} D. Rodriguez-Martin, A. Sama, C. Perez-Lopez, A. Catala, J. Cabestany, and A. Rodriguez-Molinero, “Svm-based posture identification with a single waist-located triaxial accelerometer,” Expert Systems with Applications, vol. 40, no. 18, pp. 7203–7211, 2013.
\bibitem{b20} A. Sama, D. Rodriguez-Martin, C. Perez-Lopez, A. Catala, S. Alcaine, B. Mestre, A. Prats, M. C. Crespo, and A. Bayes, “Determining the optimal features in freezing of gait detection through a single waist accelerometer in home environments,” Pattern Recognition Letters, vol. 105, pp. 135–143, 2018.
\bibitem{b21} M. Mancini and F. B. Horak, “Potential of apdm mobility lab for the monitoring of the progression of parkinson’s disease,” Expert review of medical devices, vol. 13, no. 5, pp. 455–462, 2016.
\bibitem{b22} S. T. Moore, H. G. MacDougall, and W. G. Ondo, “Ambulatory monitoring of freezing of gait in parkinson’s disease,” Journal of neuroscience methods, vol. 167, no. 2, pp. 340–348, 2008.
\bibitem{b23} D. Rodriguez-Martin, A. Sama, C. Perez-Lopez, A. Catala, J. Cabestany, and A. Rodriguez-Molinero, “Svm-based posture identification with a single waist- located triaxial accelerometer,” Expert Systems with Applications, vol. 40, no. 18, pp. 7203–7211, 2013.
\bibitem{b24} J. Hannink, T. Kautz, C. F. Pasluosta, K.-G. Gaßmann, J. Klucken, and B. M. Eskofier, “Sensor-based gait parameter extraction with deep convolutional neural networks,” IEEE journal of biomedical and health informatics, vol. 21, no. 1, pp. 85–93, 2016.
\bibitem{b25} A. Amini, K. Banitsas, and W. R. Young, “Kinect4fog: Monitoring and improving mobility in people with parkinson’s using a novel system incorporating the microsoft kinect v2,” Disability and Rehabilitation: Assistive Technology, vol. 14, no. 6, pp. 566–573, 2019.
\bibitem{b26} K. Hu, Z. Wang, W. Wang, K. A. E. Martens, L. Wang, T. Tan, S. J. Lewis, and D. D. Feng, “Graph sequence recurrent neural network for vision-based freezing of gait detection,” IEEE Transactions on Image Processing, vol. 29, pp. 1890–1901, 2019.
\bibitem{b27} K. Hu, Z. Wang, S. Mei, K. A. E. Martens, T. Yao, S. J. Lewis, and D. D. Feng, “Vision-based freezing of gait detection with anatomic directed graph representation,” IEEE journal of biomedical and health informatics, vol. 24, no. 4, pp. 1215–1225, 2019.
\bibitem{b28} Addison Howard, amit salomon, eran gazit, Jeff Hausdorff, Leslie Kirsch, Maggie, Pieter Ginis, Ryan Holbrook, and Yasir F Karim. Parkinson's Freezing of Gait Prediction. https://kaggle.com/competitions/tlvmc-parkinsons-freezing-gait-prediction, 2023. Kaggle.
\bibitem{b29} K. Ziegler, F. Schroeteler, A. O. Ceballos-Baumann, and U. M. Fietzek, “A new rating instrument to assess festination and freezing gait in parkinsonian patients,” Movement Disorders, vol. 25, no. 8, pp. 1012–1018, 2010.
\bibitem{b30} Li, Wendan, et al. "Recognition of freezing of gait in parkinson’s disease based on machine vision." Frontiers in Aging Neuroscience 14 (2022): 921081.
\bibitem{b31} Salomon, Amit, et al. "A machine learning contest enhances automated freezing of gait detection and reveals time-of-day effects." Nature Communications 15.1 (2024): 4853.
\end{thebibliography}
\end{document}